\pdfoutput=1

\documentclass[11pt]{article}

\usepackage{ACL2023}

\usepackage{times}
\usepackage{latexsym}

\usepackage[T1]{fontenc}

\usepackage[utf8]{inputenc}

\usepackage{microtype}

\usepackage{inconsolata}

\title{LLMs in Education: Novel Perspectives, Challenges, and Opportunities}

\author{Bashar Alhafni$^1$, Sowmya Vajjala$^2$, Stefano Bannò$^3$, \\ {\bf Kaushal Kumar Maurya$^4$, Ekaterina Kochmar$^4$} \\
  $^1$New York University, $^2$NRC-CNRC, $^3$University of Cambridge, $^4$MBZUAI \\
  \texttt{alhafni@nyu.edu, sowmya.vajjala@nrc-cnrc.gc.ca, sb2549@eng.cam.ac.uk,}\\ \texttt{\{kaushal.maurya, ekaterina.kochmar\}@mbzuai.ac.ae} \\
}

\begin{document}

\maketitle



\begin{abstract}
The role of large language models (LLMs) in education is an increasing area of interest today, considering the new opportunities they offer for teaching, learning, and assessment. This cutting-edge tutorial provides an overview of the educational applications of NLP and the impact that the recent advances in LLMs have had on this field. We will discuss the key challenges and opportunities presented by LLMs, grounding them in the context of four major educational applications:  reading, writing, and speaking skills, and intelligent tutoring systems (ITS). This COLING 2025 tutorial is designed for researchers and practitioners interested in the educational applications of NLP and the role LLMs have to play in this area. It is the first of its kind to address this timely topic.\footnote{Website: \url{https://coling2025-edu-llms.github.io}}

\end{abstract}

\section{Introduction}


Large language models (LLMs) such as GPT-3.5 (ChatGPT),\footnote{\url{https://openai.com/index/chatgpt/}} GPT-4 \cite{gpt4}, LLaMa \cite{touvron2023llama2openfoundation}, Falcon \cite{almazrouei2023falconseriesopenlanguage}, among others, have demonstrated remarkable capabilities across various tasks~\cite{wei2022emergent,minaee2024large}. The rapid adoption of LLMs and generative AI by EdTech companies such as Duolingo~\cite{naismith2023automated} and Grammarly \cite{raheja2023coedit} shows the strong impact recent advances in LLMs are having on the development of educational applications. The importance of this topic for the NLP community is also demonstrated by the recent panel discussion at NAACL 2024, which focused on the impact of LLMs on education.\footnote{\url{https://2024.naacl.org/program/panel/}} While LLMs have undoubtedly caused a paradigm shift in educational applications research and development -- enabling new opportunities in writing assistance, personalization, and interactive teaching and learning, among other tasks -- they also present novel challenges, particularly regarding the ethical considerations in integrating LLMs into educational settings, fair assessment, and evaluation~\cite{bommasani2021opportunities}. In this tutorial, we will delve into the challenges and opportunities presented by LLMs for educational applications, examining them through four key educational tasks.


\section{Outline}


This tutorial will last 4 hours, including a 30-minute coffee break. We will begin with a 15-minute introduction, followed by four 45-minute sessions on key educational applications. We will discuss the impact of LLMs on: (1) writing assistance; (2) reading assistance; (3) spoken language learning and assessment; and (4) the development of Intelligent Tutoring Systems (ITS). The final 15 minutes will be dedicated to conclusions and discussions.





\subsection{Introduction (15 mins)} 
We will start the tutorial with a brief overview of  the field of educational applications, motivating our choice of the four specific tasks that we discuss in more detail and linking previous research to the recent advances in the field. 


\subsection{LLMs for Writing Assistance (45 mins)}
In the context of writing assistance, this tutorial will focus on the popular task of Grammatical Error Correction (GEC), which involves automatically detecting and correcting errors in text, offering significant pedagogical benefits to both native (L1) and foreign (L2) language teachers and learners. These benefits include providing instant feedback on writing and supporting personalized learning by profiling users' grammatical knowledge.

\paragraph{\bf Overview (15 mins):}
First, we will provide an overview of the long history of GEC in computational linguistics and the recent progress witnessed over the last decade \cite{bryant:2023:survey}, introducing the most popular GEC datasets for English and other languages \citep[\textit{inter alia}]{yannakoudakis-etal-2011-new,napoles-etal-2017-jfleg,naplava-etal-2022-czech}. We will then describe different evaluation methods and their reliability~\cite{dahlmeier-ng-2012-better,felice-briscoe-2015-towards,grundkiewicz-etal-2015-human}, and summarize various GEC techniques from rule-based to sequence-to-sequence to text-editing neural models~\cite{kochmar-etal-2012-hoo,alhafni-etal-2023-advancements,omelianchuk-etal-2020-gector}.

\paragraph{\bf LLMs for GEC (15 mins):} We will provide a thorough overview on using LLMs for GEC, including an assessment of their overall performance in terms of fluency, coherence, and the ability of fixing different error types across popular GEC benchmarks \citep[\textit{inter alia}]{fang2023chatgpthighlyfluentgrammatical,raheja-etal-2024-medit,katinskaia-yangarber-2024-gpt-3}. We will delve into different prompting techniques and strategies for GEC, evaluating their effectiveness and limitations~\cite{loem-etal-2023-exploring}, and compare LLMs to supervised GEC approaches, examining their respective strengths and weaknesses~\cite{omelianchuk-etal-2024-pillars}. Finally, we will discuss how LLMs could be used to evaluate GEC systems \cite{kobayashi-etal-2024-large} and provide explanations \cite{lopez-cortez-etal-2024-gmeg-exp,kaneko-okazaki-2024-controlled-generation}
%

\paragraph{\bf Future Directions and Discussion (15 mins):} We will provide insights into future directions of using LLMs for GEC in terms of evaluation, usability, and interpretability from a user-centric perspective.

\subsection{LLMs for Reading Assistance (45 mins)}
Reading is an essential part of literacy, and the ability to read and comprehend text plays a major role in critical thinking and effective communication. Over the past two decades, NLP research has focused on two topics related to this aspect -- readability assessment and text simplification. In this section, we will give an overview of these two topics, discuss the role of LLMs in advancing research in this direction, and end with a summary of the state of the art, its limitations, and future directions. 

\paragraph{\bf Overview (15 mins):} We will introduce the tasks of readability assessment \cite{vajjala2022trends} and text simplification \cite{alva2020data}, give an overview of the different approaches explored in NLP research on this topic over time, and the challenges involved. We will also cover research that focuses on domain specific issues \cite{garimella-etal-2021-domain} as well as multilingual approaches \cite{saggion2022findings}. 

\paragraph{\bf LLMs for Reading Support (15 mins): } Continuing from the overview, we will discuss how the arrival of LLMs has led to new advances in these two tasks and how these differ from the previous state of the art. Starting from zero-shot usage of LLMs for these tasks \cite{kew2023bless}, we will summarize research into prompt-tuning and fine-tuning of LLMs \cite{lee2023prompt} and personalization \cite{tran2024readctrl} in the context of readability assessment and text simplification. 

\paragraph{\bf Current Limitations and Future Directions (15 mins):} In the final section, we will elaborate on the limitations of current approaches \cite{vstajner2021automatic}, summarize some recent trends such as the increased focus on user-based evaluation \cite{vajjala2019understanding, sauberli2024digital,agrawal2024text}, and supporting more languages \cite{shardlow-etal-2024-bea}. We will also summarize and identify impactful future directions for research in this area. 

\subsection{LLMs for Spoken Language Learning and Assessment (45 mins)} 
Speaking is a crucial language skill, which is at the core of language education curricula~\citep{fulcher2000}. There has been an increasing interest in developing methods to automate spoken language proficiency assessment, and this tutorial will focus on automatic L2 speaking assessment and feedback. In addition to holistic assessment, we will also discuss analytic assessment breaking down a response into specific components and assigning separate scores to each of them, as well as feedback with a particular focus on spoken grammatical error detection (GED) and correction (GEC).


\paragraph{\bf Overview (15 mins):} We will start with an overview of the history of automated speaking assessment~\citep{zechner2019automated} and, since many approaches to speaking assessment are borrowed from NLP, a brief mention of its counterpart -- automatic writing assessment -- overviewing the most significant studies on automatic essay scoring from the 1990s-2000s~\citep{burstein2002erater, rudner2006evaluation, landauer2002intell} and the first experimental works on pronunciation assessment~\citep{bernstein1990automatic, cucchiarini1997automatic, franco2000sri}. We will touch upon the breakthrough of the first commercial systems for speech assessment~\citep{townshend1998estimation, xi2008speechrater}, and then focus on deep neural network approaches both for writing~\citep{alikaniotis2016automatic} and speaking~\citep{malinin2017} assessment. Finally, we will talk about the applications of analytic assessment and spoken GE~\citep{Knill2019,lu2020spoken,banno2022view}.


\paragraph{\bf LLMs for Speaking Assessment and Feedback (15 mins):} 
We will then focus on the use of text-based foundation models, e.g., BERT~\citep{devlin2019bert}, and speech foundation models, e.g., wav2vec 2.0~\citep{baevski2020wav}, HuBERT~\citep{hsu2021hubert}, and Whisper~\citep{radford2022robust}. The former have been investigated for holistic assessment~\citep{craighead2020, raina2020universal, wang2021}, and the latter -- for mispronunciation detection and diagnosis~\citep{peng2021mispron}, pronunciation assessment~\citep{kim2022pron}, and analytic and holistic assessment~\citep{banno2023proficiency, banno2023slate}. We will address the issues of interpretability, exploring analytic assessment approaches for writing that are applicable to spoken transcriptions~\cite{banno2024gpt}. Additionally, we will discuss an end-to-end approach based on Whisper for spoken GEC~\cite{banno2024towards}.


\paragraph{\bf Future Directions and Discussion (15 mins):} We will explore opportunities for using LLMs in spoken language assessment, with a particular emphasis on multimodal models like SALMONN~\cite{tang2024salmonn} and Qwen Audio~\cite{qwenaudio2023}, and examine how text-to-speech models like Bark~\citep{schumacher2023enhancing} and Voicecraft~\cite{peng2024voicecraft} can support assessment and learning.


\subsection{LLMs in ITS (45 mins)}

Intelligent tutoring systems (ITS) are computerized learning environments that incorporate computational models and provide feedback based on learning progress \cite{graesser2001intelligent}. LLM-powered ITS are capable of providing learners with one-on-one tutoring, enabling equitable and pedagogically sound learning experience, which has been shown to lead to substantial learning gains~\cite{bloom19842}. This tutorial will provide an overview of the extensive literature on ITS, and then focus on advances in ITS powered by LLMs.


\paragraph{\bf Overview (15 mins):}
We will begin by outlining how lack of individualized tutoring leads to less effective learning and dissatisfaction~\cite{brinton2014individualization, eom2006determinants, hone2016exploring}, especially in big classrooms. We will then review key principles of learning sciences \cite{craighead2004concise, weinstein2018teaching}, setting the goals for ITS development, and provide a brief overview of pre-LLM ITS~\cite{paladines2020systematic}, including systems tailored for misconception identification \cite{graesser1999autotutor,rus2013recent}, model-tracing tutors \cite{rickel2002collaborative, heffernan2008expanding}, constraint-based models \cite{mitrovic2005effect}, Bayesian network models \cite{pon2004evaluating}, and systems designed to cater to specific knowledge areas \cite{weerasinghe2006facilitating} and educational levels. Most of the systems discussed were primarily developed for STEM subjects~\cite{kochmar2022automated}, with some exceptions including language learning \cite{slavuj2015intelligent}.

\paragraph{\bf LLMs in ITS Development (15 mins):} We will focus on the use of LLMs for various STEM educational tasks \cite{wang2024large}, including question solving \cite{cheng2024dated}, error correction \cite{amini2019mathqa}, confusion resolution \cite{balse2023evaluating}, question generation \cite{grenander2021deep,elkins2024teachers}, and content creation \cite{li2024automated}, among others. We will contextualize the aforementioned tasks to demonstrate that LLMs can simulate student interactions for teaching assistants \cite{markel2023gpteach} and teacher training using ITS \cite{wang2023chatgpt}, and they can also assist in creating datasets \cite{vasselli2023naisteacher} used for fine-tuning LLMs \cite{macina2023mathdial} and employ prompt-based techniques \cite{wang-etal-2024-bridging} or modularized prompting \cite{daheim2024stepwise} for the development of ITS.  

\paragraph{\bf Future Directions and Discussion (15 mins):}
Despite significant advances in LLM-driven ITS, several challenges remain. We will outline potential future directions such as: (i) the development of standardized evaluation benchmarks to assess and track the progress in ITS; (ii) the collection and creation of large public educational datasets for LLM training and fine-tuning, as existing efforts \cite{macina2023mathdial, wang-etal-2024-bridging, stasaski2020cima} are limited to a few hundred thousand examples; and (iii) the development of specialized foundational LLMs for educational purposes. We will argue for investigations into the long-term impact of LLM-powered ITS on students and teachers, while also examining ethical considerations, potential biases, and the pedagogical value of the generated content in dialogue-based ITS.




\subsection{Conclusions \& Discussion (15 mins)} 

In this final part of the tutorial, we will summarize the content covered, outline useful resources in this field, and open the floor for discussion.

\section{Reading List}

In addition to the papers cited in this proposal, we also recommend the reading list that is available on the tutorial \href{https://coling2025-edu-llms.github.io/}{website}.
Proceedings of the \href{https://aclanthology.org/sigs/sigedu/}{Workshop on Innovative Use of NLP for Building Educational Applications (BEA)} are also highly relevant.

\section{Target Audience}



The target audience includes graduate students, researchers, and practitioners attending COLING 2025 with the background in Computational Linguistics (CL), NLP, and/or Machine Learning (ML), interested in educational applications of NLP and generative AI. Basic knowledge of educational technologies is helpful but not required. The tutorial will be self-contained and accessible to a wide audience. We estimate 50 to 100 participants based on recent BEA Workshop attendance and growing interest in AI applications in education, as seen at the NAACL 2024 panel on {\em LLMs and their Impact on Education}.



\section{Type of the Tutorial}

The tutorial is designed to be {\bf cutting edge}, covering advanced technologies for a range of educational applications. Specifically, we will focus on the recent advances brought to the field by generative AI and LLMs, discussing new opportunities as well as new challenges and risks. To the best of our knowledge, this is the first tutorial on this topic organized at COLING or any other *CL conference.

\section{Diversity Considerations}


Our tutorial covers a wide range of applications, including seminal and recent work in the field. It is designed for attendees with diverse backgrounds interested in educational applications. We believe AI can democratize education and increase accessibility worldwide. We will highlight opportunities to reach underrepresented groups and address challenges related to fairness and accessibility. The instructors are diverse in gender, nationality, affiliation, and seniority (from PhD students to postdocs to professors). The tutorial will include open Q\&A sessions for participant engagement and discussion.

\section{Presenters}



\paragraph{\bf{\href{https://www.basharalhafni.com}{Bashar Alhafni}}} is a final-year Computer Science PhD student at New York University and a Graduate Research Assistant at the Computational Approaches to Modeling Language (CAMeL) Lab in New York University Abu Dhabi. His research focuses on controlled natural language generation tasks such as grammatical error correction, text simplification, and user-centric text rewriting. 


\paragraph{\bf{\href{https://nrc.canada.ca/en/corporate/contact-us/nrc-directory-science-professionals/sowmya-vajjala}{Sowmya Vajjala}}} is a researcher at the National Research Council, Canada. Her current research focuses on text classification and information extraction in multilingual contexts. In the past, she has extensively worked on research focusing on the educational applications of NLP such as readability and language proficiency assessments. 

\paragraph{\bf{\href{https://www.eng.cam.ac.uk/profiles/sb2549}{Stefano Bannò}}} is a Research Associate at Cambridge University’s Institute for Automated Language Teaching and Assessment (ALTA) and the Machine Intelligence Lab, Department of Engineering. His research interests span automatic assessment of L2 spoken English, exploring learner-oriented feedback, analytic assessment of individual proficiency aspects, and spoken grammatical error correction.
\paragraph{\bf{\href{https://kaushal0494.github.io}{Kaushal Kumar Maurya}}} is a post-doctoral associate in the NLP Department at MBZUAI. His research focuses on advancing research in intelligent tutoring systems powered by LLMs. His work includes developing evaluation benchmarks, datasets, systems, and assessment tools/frameworks.  


\paragraph{\bf{\href{https://ekochmar.github.io/about/}{Ekaterina Kochmar}}} is an Assistant Professor at the NLP Department at MBZUAI, where she conducts research at the intersection of AI, NLP, and ITS. Her research on developing educational applications for L2 learners contributed to building {\em Read \& Improve}, a readability platform for non-native English readers. Her current research on AI-powered dialogue-based ITS resulted in the development of {\em Korbit AI} tutor, a system capable of providing learners with high-quality and personalized education.

\section*{Ethics Statement}
This tutorial provides a comprehensive overview of educational applications, focusing on novel perspectives, opportunities, and challenges in the era of LLMs, using examples from four key educational tasks. Education is a high-impact area for integrating LLMs into real-life applications. Thus, we believe that it is important to discuss the impact of new technologies on education and raise awareness of potential risks. Each section of the tutorial includes a discussion on the ethical considerations and potential risks of using LLMs within the respective application context.

\bibliography{anthology,custom}
\bibliographystyle{acl_natbib}

\end{document}